\documentclass[10pt,conference]{IEEEtran}
\makeatletter
\IEEEoverridecommandlockouts

\def\ps@headings{%
\def\@oddhead{\mbox{}\scriptsize\rightmark \hfil \thepage}%
\def\@evenhead{\scriptsize\thepage \hfil \leftmark\mbox{}}%
\def\@oddfoot{}%
\def\@evenfoot{}}
\makeatother 

\usepackage{color}
\usepackage{graphicx}               
\usepackage[cmex10]{amsmath}

\ifCLASSINFOpdf
\else
\fi

\usepackage{algorithm}

\usepackage{algpseudocode}
\usepackage{caption}

\usepackage{url}

\begin{document}
%
\title{Demo: Edge-centric Telepresence Avatar Robot for Geographically Distributed Environment}

\author{\IEEEauthorblockN{Ashis Sau, Ruddra Dev Roychoudhury, Hrishav Bakul Barua, Chayan Sarkar, Sayan Paul,\\ Brojeshwar Bhowmick, Arpan Pal, and Balamuralidhar P}
\IEEEauthorblockA{TCS Research \& Innovation, India}
}


%


\maketitle



%
\IEEEpeerreviewmaketitle

\begin{abstract}
	Using a robotic platform for telepresence applications has gained paramount importance in this decade. Scenarios such as remote meetings, group discussions, and presentations/talks in seminars and conferences get much attention in this regard. Though there exist some robotic platforms for such telepresence applications, they lack efficacy in communication and interaction between the remote person and the avatar robot deployed in another geographic location. Also, such existing systems are often cloud-centric which adds to its network overhead woes. In this demo, we develop and test a framework that brings the best of both cloud and edge-centric systems together along with a newly designed communication protocol. Our solution adds to the improvement of the existing systems in terms of robustness and efficacy in communication for a geographically distributed environment.    
\end{abstract}

\section{Introduction}
We are fast approaching an era of unprecedented success in robotics. Robots are being deployed in many scenarios nowadays and telepresence is one of the most prominent applications across varied applications. A teleoperated robot finds application in almost all sectors such as industry, academia, healthcare, government, etc. The most prominent application considers representing a remote person in a meeting, discussion, seminar or conference being his avatar \cite{rae2017robotic}. Such an avatar robot needs a very robust underlying framework to work with efficacy~\cite{wang2015interfacing}. The current state-of-the-art lacks a robust communication and interactive framework for such an avatar robot\footnote{\url{https://telepresencerobots.com/robots/ava-robotics-ava-500}}. Moreover, the existing systems for such telepresence robots are cloud-centric and so it also introduces some communication delays, which hampers many real-time applications to a great extent. So, we take this problem and design a robust edge-centric communication framework to improve the state-of-the-art systems.            

We develop a telepresence solution using the Double-2 robot, a state-of-the-art telepresence robotic platform\footnote{\url{https://www.doublerobotics.com/double2.html}}. The platform provider uses a resource-constrained device (an Apple iPad) to interface the hardware and hosts all the computation on the cloud. We develop an edge-centric model to create a navigation and interaction solution. The edge-centric model supports robust real-time communication between the edge and the avatar as well as the master, the user of the robotic platform. We develop an application layer protocol that caters to the need of the navigation and interaction framework with provisions for extensions in the future.


\begin{figure*}[]
	\centering
	\includegraphics[width=\linewidth]{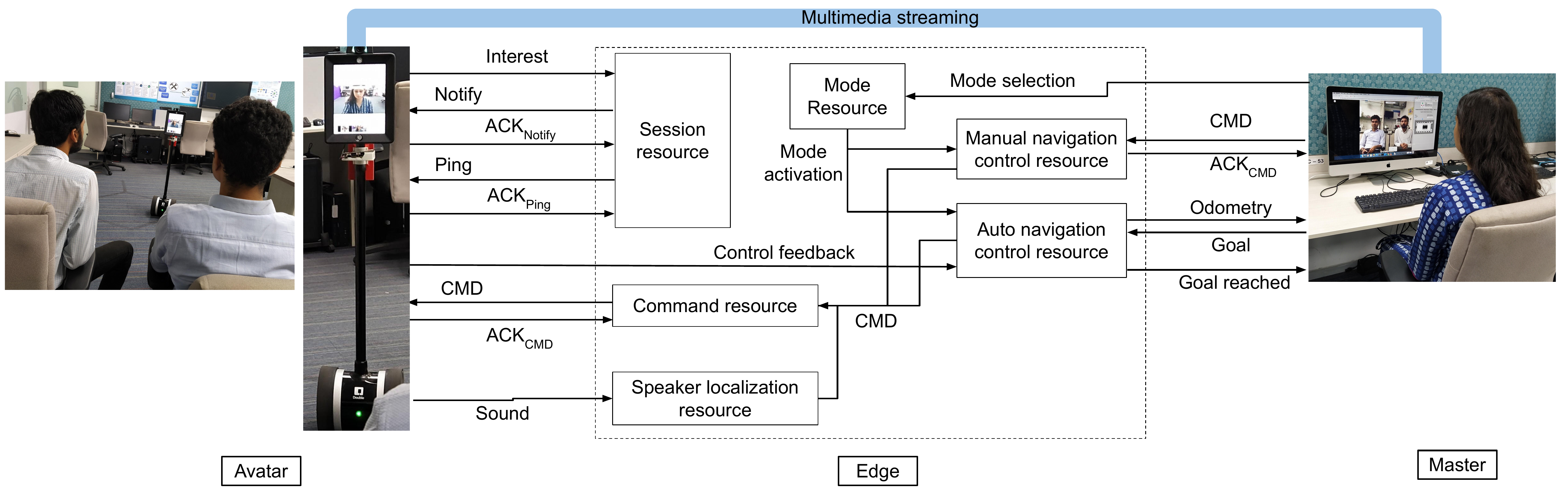}
	\caption{Architecture of a distributed edge-centric telepresence system using a mobile robot.}
	\label{fig:architecture}
\end{figure*}

When high-level task planning and interaction for a robot is delegated to an external computing resource (edge computing device), there must be a low latency but highly reliable communication protocol that can transfer a variety of control commands to the robot avatar. It should also be able to receive and interpret the status messages sent back from the robot for effective seamless communication. Since Double-2 operates on iOS, we develop a new framework that interfaces between Robotic Operating System (ROS) and iOS efficiently. This framework has the inherent advantage of not being cloud-centric and able to handle the security requirements. It can also be modeled for other large networks possibly in universities and corporate organizations where this framework is planned to be deployed. Apart from this, we also create provisions of connection between avatar and master (remote user) using Cisco WebEx technologies. WebEx provides a video conferencing interface for the master, where the WebEx runs in the foreground and the communication and interaction framework runs in the back-end of the iPad attached to the robot. 

\section{Architecture and development of the framework}
Figure~\ref{fig:architecture} depicts the overall architecture of our edge-centric telepresence framework and its various modules. The avatar (robotic platform) can be controlled in two modes -- manual and auto mode. In the case of remote navigation, the master can choose and alter his/her navigation mode preference anytime during the usage. The selected mode is saved in mode resource (residing at the edge device), which triggers the selected navigation control module accordingly. Then, the particular control module sends commands to the avatar through the command resource. In manual mode, the master instructs and controls the robot using low-level commands like park, turn-left, turn-right, drive-left, drive-right, stop-drive, etc. On the other hand, in auto mode, one map of the meeting room is shown to master with the avatar's current position. Master can send the destination goal by clicking at any point inside the map. The path planner in the edge device processes the input and generates the coordinates of the waypoints to reach the destination. The determined path breaks down into small paths by a path-discretizer module and later it generates and sends low-level command sequences to the avatar for actuation. This path-discretizer helps in more correct navigation. During this actuation, the master gets informed with the avatar's current position by sending the current odometry from the edge at a predefined interval. 

The framework contains a session resource, which is used to continuously exchange ping messages (during the ideal period) between Edge and Avatar to keep the connection alive. The Command resource receives commands from the Autonomous and the Manual control resources and translates them to the robot in a synchronous manner i.e., a command is sent to the Avatar only when it receives the acknowledgment that the previous command has completed execution and the Avatar is connected to the Edge. 
In our telepresence system two different aspects of an agile meeting scenario, robot maneuvering, and multimedia communication, are maintained parallelly but in a synchronized manner. Though the robot maneuvering is designed to be controlled by the master from a different geographic location, it is designed to be implicitly controlled by the local attendees as well to make the meeting agile. We introduce a speaker localization-based attention shifting mechanism so that the co-located people get a feeling of attentiveness from the avatar's perspective. One raspberry pi with 4-mic-array has also been integrated with the robot. It senses the person speaking in the meeting and finds the angle of the speaking person with respect to its current pose. After that, it sends the angle to edge which is processed in the speaker-localization resource to generate the specific turning command so that the avatar can turn towards the speaker ~\cite{barua2019tele}.

For communication between the edge-centric computing device and avatar, we design a UDP-based low-latency application-level protocol. UDP provides an advantage in real-time communication by making communication in a best-effort way.  As our protocol is designed over UDP, we design application-level reliability by sending acknowledgments to edge whenever required as the idea has been introduced in A-REaLiST \cite{bhattacharyya2018improving}. Edge maintains a status queue for commands in command resource and master also gets informed about the status of the command it sends. During auto navigation, resending of the lost command sequence mechanism is also incorporated from the edge in a certain interval. For communication between the master and the edge, we use HTTP web-service based API. 

We use a ROS enabled Ubuntu system as Edge and Double-2 robot as an avatar. We build our custom iOS application for the iPad on Double-2 which is using a double-SDK to actuate with hardware in Double-2. Master can connect using a URL in Chrome or Firefox browser. Multimedia communication is established parallelly using standard Cisco WebEx setup. In the iPad integrated with Double-2, our custom application runs in background mode whereas the cisco WebEx meeting application runs in the foreground.

\section{Conclusion}
Our system displays a high degree of usability and reproducibility for its simple yet robust design and can be used for a meeting as well as other group discussion or educational applications such as giving proxy of a student in school if he/she is sick. The current system does not have NLP subsystems for understanding natural language commands given by a remote person or co-located people~\cite{pramanick2019enabling}. We wish to integrate such sub-systems into the Double-2 framework in the future.

\bibliographystyle{IEEEtran}
\bibliography{main}
\end{document}